\documentstyle[myapa,latex-acl]{article}
\input{psfig}

\newcommand{\fs}[1]
  {\vspace*{0.1cm}
    \mbox{\small%
          $
          \left[
          \!\!
          \it
          \begin{tabular}{ll}
          #1
          \end{tabular}
          \!\!
          \right]
          $}
   }

\begin{document}

\pagestyle{empty}

\title{\vspace{-0.5in}\LARGE\bf VERB SEMANTICS AND LEXICAL SELECTION}
\author{
Zhibiao Wu \\
Department of Information System\\ \& Computer Science \\
National University of Singapore \\
Republic of Singapore, 0511 \\
wuzhibia@iscs.nus.sg \And
Martha Palmer \\
Department of Computer and\\
Information Science \\
University of Pennsylvania \\
Philadelphia, PA 19104-6389 \\
mpalmer@linc.cis.upenn.edu}

\maketitle
\vspace{-0.5in}

\begin{abstract}
This paper will focus on the semantic representation of verbs
in computer systems and its impact on lexical selection problems in
machine translation (MT). Two groups of English and Chinese verbs are
examined to show that lexical selection must be based on interpretation
of the sentence as well as selection restrictions placed on
the verb arguments. A novel representation scheme is suggested,
and is compared to representations with selection restrictions
used in transfer-based MT.  We see our approach as closely aligned
with knowledge-based MT approaches (KBMT), and as a separate component
that could be incorporated into existing systems.
Examples and experimental results will show that, using this scheme,
inexact matches can achieve correct lexical selection.
\end{abstract}

\section*{Introduction}
The task of lexical selection in machine translation (MT)
is choosing the target lexical item which
most closely carries
the same meaning as the corresponding item in the source text.
Information sources that support this decision
making process are the source text, dictionaries, and knowledge
bases in MT systems. In the early direct replacement approaches,
very little data was used for verb selection.
The source verb
was directly replaced by a target verb with the help of
a bilingual dictionary. In transfer-based approaches,
more information is involved
in the verb selection process. In particular,
the verb argument structure is used for selecting the target
verb. This requires that each translation verb pair and
the selection restrictions on the verb arguments be exhaustively
listed in the bilingual dictionary. In this way, a verb sense is
defined with a target verb and
a set of selection restrictions on its arguments.
Our questions are: Is the exhaustive listing of translation verb pairs
feasible? Is this verb representation scheme sufficient
for solving the verb selection
problem? Our study of a particular MT system shows that when
English verbs are translated into Chinese,
it is difficult to achieve large coverage by listing translation pairs.
We will show that a set of
rigid selection
restrictions on verb arguments
can at best define a default situation for the verb usage.
The translations from English verbs to Chinese verb compounds that we
present here provide
evidence of the reference to the context and to a fine-grained level
of semantic representation.
Therefore, we propose a novel verb
semantic representation that defines each verb by a set of concepts in
different conceptual
domains. Based on this conceptual representation, a similarity measure
can be defined that allows correct lexical choice to be achieved, even
when there is no exact lexical match from the source language to the
target language.

We see this approach as compatible with other
interlingua verb representation methods, such as verb representations
in KBMT (Nirenburg,1992) and UNITRAN \cite{dorr:lexical}.
Since these
methods do not currently
employ a multi-domain approach, they cannot address the
fine-tuned meaning differences among verbs and the correspondence between
semantics and syntax.  Our
approach could be adapted to either of these
systems and incoporated into them.

\section*{The limitations of direct transfer}

In a transfer-based MT system, pairs of verbs are exhaustively listed in a
bilingual dictionary. The translation of a source verb is limited
by the number of entries in the dictionary.
For some source verbs with just a few translations, this method
is direct and efficient. However, some source verbs are very active and
have a lot of different translations in  the target language.
As illustrated by the following test of a
commercial English to Chinese MT system, TranStar,
using sentences from the Brown corpus,
current transfer-based approaches have no alternative to
 listing every translation pair.

In the Brown corpus, 246 sentences take {\it break} as the main verb.
After removing most idiomatic usages and verb particle
constructions, there are
157 sentences left. We used these sentences to test TranStar.
The translation results are shown
below:

\centerline{\tiny
\begin{tabular}{lll}
\psfig{figure=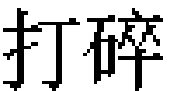,height=8pt}   107
& \psfig{figure=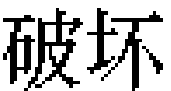,height=8pt}   22
&  \psfig{figure=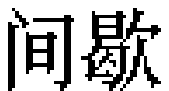,height=8pt}   14 \\
dasui & pohui & jianxie \\
to break into pieces &
to make damage to & to have a break \\
 \psfig{figure=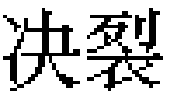,height=8pt}  5
&  \psfig{figure=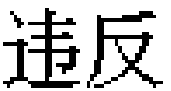,height=8pt}  2
&  \psfig{figure=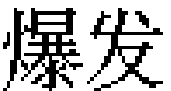,height=8pt}  0 \\
juelie & weifan & baofa \\
 to break (a relation) &
to against & to break out \\
 \psfig{figure=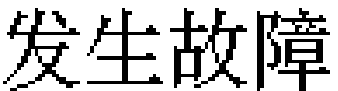,height=8pt}  0
&  \psfig{figure=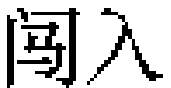,height=8pt} 0
&  \psfig{figure=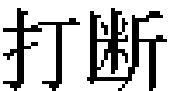,height=8pt}  0 \\
fashenguzhang & chuanlu & daduan \\
 to break down & to break into &
to break a continuity \\
  \psfig{figure=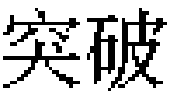,height=8pt} 0
&  \psfig{figure=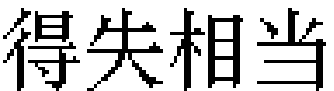,height=8pt}  0
&   \psfig{figure=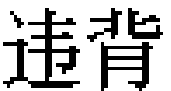,height=8pt}  0 \\
tupo & deshixiandan & weibei \\
 to break through & to break even with
& to break (a promise) \\
 \psfig{figure=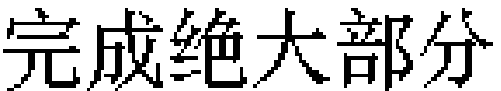,height=8pt} 0 \\
wanchenjuedabufen \\
to break with
\end{tabular} }

In the TranStar system,  English {\it break} only has 13 Chinese verb
entries. The numbers above are the frequencies with which
the 157 sentences translated into a particular Chinese expression.
Most of the zero frequencies represent Chinese verbs that correspond to
English {\it break}
idiomatic usages or verb particle constructions
which were removed.
The
accuracy rate of the translation is
not high. Only 30 (19.1\%) words were correctly translated.
The Chinese verb \psfig{figure=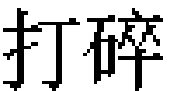,height=8pt} (dasui) acts like a
default
translation when no other choice matches.

The same 157 sentences were translated by one of the authors into
68 Chinese verb expressions.  These expressions can be listed
according to the frequency with which they occurred, in decreasing
order.
The verb which has the highest rank is the verb which has the
highest frequency. In this way, the frequency distribution of the two
different translations can be shown below:

\centerline{\psfig{figure=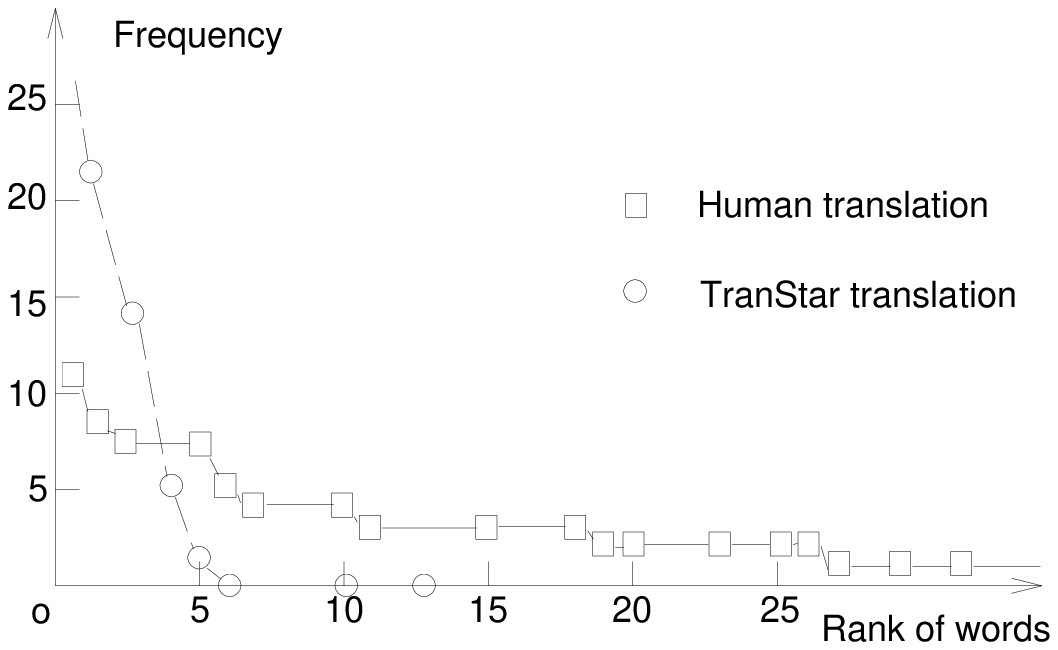,height=1in,width=2in}}
\centerline{\small Figure 1. Frequency distribution of translations}

It seems that the nature of the lexical selection task in translation
obeys Zipf's law. It means that, for all possible verb usages,
a large portion is translated into a few target
verbs, while a small portion might be translated into many
different target verbs. Any approach that has a fixed number of
target candidate verbs  and provides
no way to measure the meaning similarity among verbs,
is not able to handle the new verb usages, i.e., the small portion outside the
dictionary coverage. However, a native speaker has
an unrestricted number of verbs for lexical selection.
By measuring the similarities among target verbs, the most similar
one can be chosen for the new verb usage.
The challenge of verb representation is
to capture the fluid nature of verb meanings that allows human
speakers to contrive new usages in every sentence.

\section*{Translating English into Chinese serial verb compounds}

Translating the English verb {\it break} into Chinese (Mandarin) poses unusual
difficulties for two reasons.  One is that in English {\it break} can
be thought of as a very general verb indicating an entire set of
breaking events that can be distinguished by the resulting state
of the object being broken.  {\it Shatter, snap, split,} etc., can all be
seen as more specialized versions of the general breaking event.
Chinese has no equivalent verb for indicating the class of breaking
events, and each usage of {\it break} has to be mapped on to a more
specialized lexical item.  This is the equivalent of having to first interpret
the English expression into its more semantically precise situation.
For instance this would probably result
in mapping, {\it John broke
the crystal vase}, and {\it John broke the stick}
onto {\it John shattered the crystal vase} and {\it John snapped the
stick}.  Also, English specializations of {\it break}
do not cover all the ways in which Chinese can express a
breaking event.

But that is only part of the difficulty in translation.
In addition to requiring more semantically precise lexemes, Mandarin also
requires a serial verb construction.  The action by which force
is exerted to violate the integrity of the object being broken
must be specified, as well as the description of the resulting
state of the broken object itself.

\noindent {\bf Serial verb compounds in Chinese - }
Chinese serial verb compounds are composed of two
Chinese characters, with the first character
being a verb, and the second
character  being a verb or adjective. The grammatical analysis can be
found in \cite{wu:serial}. The following is an example:

\centerline{
\begin{tabular}{llll}
\psfig{figure=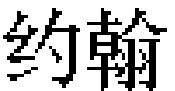,height=8pt} & \psfig{figure=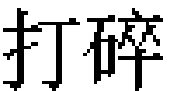,height=8pt}
& \psfig{figure=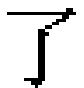,height=8pt} &
\psfig{figure=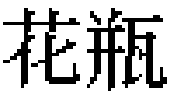,height=8pt} \\
Yuehan & da-sui & le & huapin. \\
John & hit-broken & Asp. & vase. \\
\end{tabular}}
\centerline{John broke the vase. (VA) }

Here, {\it da} is the action of John, {\it sui} is the resulting
state of the vase after
the action. These two Chinese characters are composed to form a verb
compound.
Chinese verb compounds are productive. Different verbs and adjectives
can be composed to form new verb compounds,
as in \psfig{figure=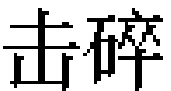,height=8pt}, ji-sui, hit-being-in-pieces; or
\psfig{figure=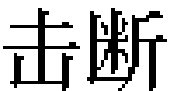,height=8pt}, ji-duan, hit-being-in-line-shape.
Many of these verb compounds have not been listed in the human
dictionary.
However, they must still be listed individually
in a machine dictionary.
Not any single character verb or single character adjective
can be composed to form a VA type verb compound. The
productive applications must
be semantically sound, and therefore have to treated individually.

\noindent {\bf Inadequacy of selection restrictions for choosing actions - }
By looking at specific examples, it soon becomes clear that
shallow selection restrictions give very little information
about the choice of the action.  An understanding of
the context is necessary.

For the sentence {\it John broke the vase}, a correct
translation is:

\centerline{
\begin{tabular}{lllll}
\psfig{figure=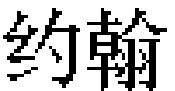,height=8pt} & \psfig{figure=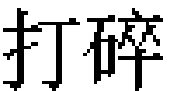,height=8pt}
& \psfig{figure=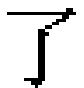,height=8pt} &
\psfig{figure=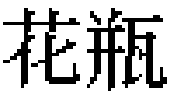,height=8pt}.  \\
 Yuehan & da-sui & le & huapin. \\
 John & hit-in-pieces & Asp. & vase. \\
\end{tabular}}

Here {\it break} is translated into a VA type verb compound.
The action is specified clearly in the translation sentence.
The following sentences which do not specify the action clearly
are anomalous.

\centerline{
\begin{tabular}{lllll}
$*$ \psfig{figure=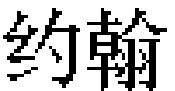,height=8pt} &
\psfig{figure=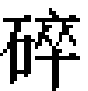,height=8pt} & \psfig{figure=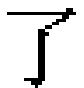,height=8pt}
&
    \psfig{figure=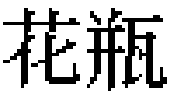,height=8pt}. \\
 Yuehan & sui & le & huapin. \\
 John & in-pieces & Asp. & vase. \\
\end{tabular}}

A translation with a causation verb is also anomalous:\\
\centerline{
\begin{tabular}{llllll}
$*$  \psfig{figure=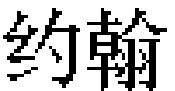,height=8pt} &
\psfig{figure=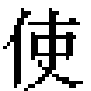,height=8pt} & \psfig{figure=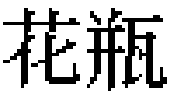,height=8pt}
&
     \psfig{figure=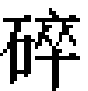,height=8pt} &
\psfig{figure=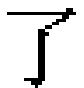,height=8pt}.\\
 Yuehan & shi & huapin & sui & le. \\
 John & let & vase & in-pieces & Asp. \\
\end{tabular}}

The following example shows that the translation must depend on
an understanding of the surrounding context.
\begin{quote}\em
 The earthquake shook the room violently, and the more fragile
 pieces did not hold up well.  The dishes shattered, and
 the glass table was smashed into many pieces.
\end{quote}
Translation of last clause:\\
\centerline{\tiny
\begin{tabular}{lllllll}
\psfig{figure=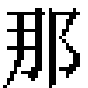,height=8pt} &
\psfig{figure=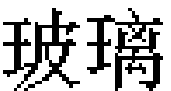,height=8pt}  &
\psfig{figure=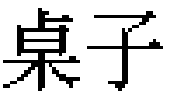,height=8pt} & \psfig{figure=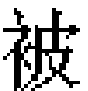,height=8pt}
& \psfig{figure=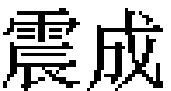,height=8pt} &
\psfig{figure=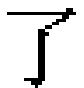,height=8pt} & \psfig{figure=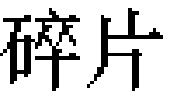,height=8pt}
\\
       na &  boli & zhuozi & bei & zhenchen &  le  &   suipian \\
       That & glass & table & Pass. &  shake-become & Asp. & pieces
\end{tabular}}

\noindent {\bf Selection restrictions reliably choose result states -}
Selection restrictions are more reliable when they
are used for specifying the
result state. For example, {\it break} in
{\it the vase broke} is translated
into {\it dasui} (hit and broken into pieces), since the vase is
brittle and easily broken into pieces.
{\it Break} in {\it the stick broke} is translated into {\it zheduan}
(bend and separated
into line-segment shape) which is a default situation for breaking
a line-segment shape object.
However, even here, sometimes the context can override the selection
restrictions on a particular noun. In {\it John broke the stick into
pieces}, the obvious translation would be {\it da sui} instead.
These examples illustrate that achieving correct lexical
choice requires more than a simple matching of selection restrictions.
A fine-grained semantic representation  of the interpretation of
the entire sentence is required.  This can indicate the contextually
implied action as well as the resulting state of the object involved.
An explicit representation of the context is beyond the state of
the art for current machine translation. When the context is
not available,
We need an algorithm for selecting the action verb.
Following is a decision tree for translating English Change-of-state
verbs into Chinese:

\centerline{\psfig{figure=action.ps,height=2.5in,width=3in}}
\centerline{\small Figure 2. Decision tree for translation}


\section*{A multi-domain approach}

We suggest that to achieve accurate lexical selection, it
is necessary to have fine-grained selection restrictions that can be
matched in a flexible fashion, and which can be augmented when necessary
by context-dependent knowledge-based understanding. The underlying
framework for both the selection restrictions on the verb arguments and
the knowledge base should be a verb taxonomy that relates verbs with
similar meanings by associating them with the same conceptual domains.

We view a verb meaning as a lexicalized concept which is undecomposable.
However, this semantic form can be projected onto a set of concepts in
different conceptual domains.  Langacker \cite{langacker:overview}
presents a set of basic domains used for defining a {\it knife}. It is
possible to
define an entity by using the {\it size, shape, color, weight,
functionality} etc.
We think it is also possible to identify a compatible set of conceptual domains
for characterizing events and therefore, defining verbs as well.  Initially
we are relying on the semantic domains suggested by Levin
as relevant to syntactic alternations, such as {\it motion},
{\it force}, {\it contact}, {\it change-of-state} and {\it action,} etc,
\cite{levin:english}.
We will augment these domains as needed  to distinguish between different
senses for the achievment of accurate lexical selection.

If words can be defined with concepts in a hierarchical structure,
it is possible to measure the meaning similarity between words with
an information measure based on WordNet \cite{resnik:selection}, or
structure level information based on a thesaurus \cite{kurohashi:dynamic}.
However, verb meanings are difficult to organize in a hierarchical structure.
One reason is that many verb meanings are involved in several
different conceptual domains. For example, {\it break} identifies a
{\it change-of-state} event with
an optional {\it causation} conception, while {\it hit}
identifies a complex event involving {\it motion}, {\it force} and {\it
contact} domains. Those Chinese verb compounds with $V + A$ constructions
always identify complex events which involve {\it action} and {\it
change-of-state}
domains.
Levin has demonstrated that in English a  verb's syntactic behavior has a
close relation to semantic components of the verb.
Our lexical selection study shows that these semantic domains are also
important for accurate lexical selection.
For example, in the above decision tree
for action selection,
a Chinese verb compound {\it dasui} can be defined with a concept {\it
\%hit-action}
in an {\it action} domain and a concept {\it \%separate-into-pieces}
in a {\it change-of-state} domain. The {\it action} domain can be further
divided into {\it motion}, {\it force}, {\it contact} domains, etc.
A related discussion
about defining complex concepts with simple concepts
can be found in \cite{ravin:lexical}.
The semantic relations of verbs that are relevant
to syntactic  behavior and that capture part of the
similarity between verbs
can be more closely realized with a conceptual
multi-domain
approach than with a paraphrase approach. Therefore we propose the following
representation method for verbs, which makes use of several different
concept domains for verb representation.

\noindent {\bf Defining verb projections - }
Following is a representation of a {\it break} sense.\\
\centerline{\tiny
\fs{LEXEME & BREAK-1 \\
EXAMPLE & I dropped my cup and it broke.\\
CONSTRAINT & (is-a physical-object E1) \\
& (is-a animate-object E0)\\
 & (is-a instrument E2) }}
\centerline{\tiny
\fs{
OBL & \fs{ch-of-state & (\%change-of-integrity E1) } \\
                  OPT & \fs{  causation & (\%cause E0 $*$) \\
  				  instrument &  (\%with-instrument E0 E2)} \\
	          IMP & \fs {      time & (\%around-time @t0 $*$)\\
 			               space &  (\%at-location @l0 E0) \\
					 &  (\%at-location @l1 E1)\\
					 &  (\%at-location @l2 E2)\\
			              action & @ \\
			       functionality & @ }}}

The CONSTRAINT slot encodes the selection information on verb arguments,
but the meaning itself is not a paraphrase.
The meaning representation is divided into three parts. It identifies a
\%change-of-integrity concept in the change-of-state domain which
is OBLIGATORY
to the verb meaning. The {\it causation} and
{\it instrument} domains are OPTIONAL and may be realized by
syntactic alternations.
Other {\it time}, {\it space}, {\it action} and {\it functionality} domains
are IMPLICIT, and are necessary for
all events of this type.

In each conceptual domain, lexicalized concepts can be organized in a
hierarchical structure. The conceptual domains for English and Chinese are
merged to form interlingua conceptual domains used for similarity
measures. Following is part of
the {\it change-of-state} domain containing
English and Chinese lexicalized concepts.

\centerline{\psfig{figure=cos.ps,height=3in,width=3in}}
\centerline{\small Figure 3. Change-of-state domain for English and Chinese}

Within one conceptual domain, the similarity of two concepts is defined
by how closely they are related in the hierarchy, i.e.,
their structural relations.

\centerline{\psfig{figure=simi1.ps,height=1.5in}}
\centerline{\small Figure 4. The concept similarity measure}

The conceptual similarity between C1 and C2 is:

\centerline{$ConSim(C1, C2) = \frac{ 2 * N3 }{N1 + N2 + 2 * N3}$}

C3 is the least common superconcept of C1 and C2. N1 is the number of
nodes on the
path from C1 to C3. N2 is the number of nodes on the path from C2 to
C3. N3 is the number of nodes on the path from C3 to root.

After defining the similarity measure in one domain, the similarity
between two verb meanings, e. g, a target verb and a source verb,
can be defined as a summation of weighted
similarities between pairs of simpler concepts in each of the domains the two
verbs are projected onto.

\centerline{$WordSim(V_1, V_2) = \sum_i W_i * ConSim(C_{i,1}, C_{i,2})$}

\section*{UNICON: An implementation}

We have implemented a prototype lexical selection system UNICON where the
representations of both the
English and Chinese verbs are based on a set of shared semantic domains.
The selection information is also included in these representations,
but does not have to match exactly.
We then organize these
concepts into hierarchical structures to form an interlingua conceptual base.
The names of our concept  domain constitute the artificial language on which
an interlingua must be based, thus place us firmly in the knowledge
based understanding MT camp.  \cite{goodman:kbmt}.

The input to the system is the source verb argument structure. After
sense disambiguation, the internal sentence representation
can be formed. The system then tries to find the target verb
realization
for the internal representation. If the concepts in the representation
do not have any target verb realization, the system takes
nearby
concepts as candidates to see whether they have target verb
realizations.
If a target verb is found, an inexact match is performed with the
target verb meaning and the internal representation, with
the selection restrictions associated with
the target verb being imposed on the input arguments.
Therefore, the
system has two measurements in this inexact match. One is the
conceptual similarity of the internal representation and the target
verb meaning, and the other is the degree of satisfaction of
the selection restrictions on the verb arguments. We take
the conceptual similarity, i.e., the meaning, as
having first priority
over the selection restrictions.

\noindent {\bf A running example - }
For the English sentence {\it The branch broke}, after disambiguation,
the internal meaning representation of the sentence can be:

\centerline{\tiny
\fs{INTER-REP & sentence-1 \\
      ch-of-state & (change-of-integrity branch-1)}}

Since there is no Chinese lexicalized concept having an exact match
for the concept {\it change-of-integrity}, the system looks at the
similar concepts in the lattice around it. They are:

{\tiny \noindent
\hspace*{3em}(\%SEPARATE-IN-PIECES-STATE \\
\noindent \hspace*{3em} \%SEPARATE-IN-NEEDLE-LIKE-STATE \\
\noindent \hspace*{3em}  \%SEPARATE-IN-DUAN-STATE \\
\noindent \hspace*{3em}  \%SEPARATE-IN-PO-STATE\\
\noindent \hspace*{3em}  \%SEPARATE-IN-SHANG-STATE\\
\noindent \hspace*{3em}  \%SEPARATE-IN-FENSUI-STATE) }

For one concept \%SEPARATE-IN-DUAN-STATE, there is a set of
Chinese realizations:

\begin{itemize}\tiny
\item \mbox{\psfig{figure=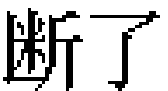,height=8pt} duan la} ( to separate
in line-segment shape).
\item
\mbox{\psfig{figure=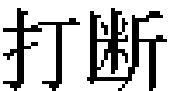,height=8pt}-1 da duan} ( to hit and separate
the object in
line-segment shape).
\item
\mbox{\psfig{figure=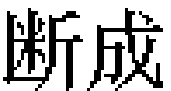,height=8pt} duan cheng} ( to separate in
line-segment
shape into).
\item
\mbox{\psfig{figure=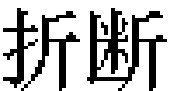,height=8pt} zhe duan} ( to bend and separate
in
line-segment shape with human hands)
\item
\mbox{\psfig{figure=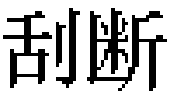,height=8pt} gua duan} ( to separate in
line-segment shape by
wind blowing).
\end{itemize}

After filling the argument of each verb representation and doing an
inexact match with the internal representation, the result is as follows:

\centerline{
\begin{tabular}{cccccc}
 & \psfig{figure=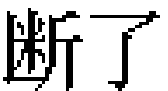,height=8pt} &
\psfig{figure=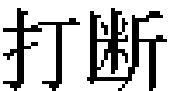,height=8pt}-1 &
\psfig{figure=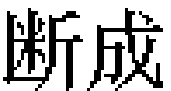,height=8pt} &
\psfig{figure=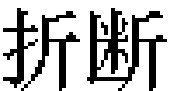,height=8pt} & \psfig{figure=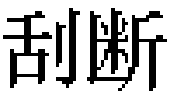,height=8pt}
\\
conceptions     & 6/7  & 0  & 0   & 0    & 0  \\
constraints & 3/14 & 0  & 3/7  & 0    & 0
\end{tabular}}

The system then chooses the verb \psfig{figure=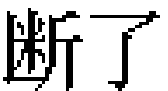,height=8pt} (duan
la) as the target
realization.

\noindent {\bf Handling metaphorical usages - }
One test of our approach was its ability to match metaphorical
usages, relying on a handcrafted ontology for the objects involved.
We include it here to illustrate the flexibility
and power of the similarity measure for handling new usages.
In these examples the system
effectively performs coercion of the verb arguments \cite{hobbs:overview}.

The system was able to translate the following metaphorical usage
from the Brown corpus correctly.
\begin{quote}\it
cf09:86:No believer in the traditional devotion of royal servitors, the plump
Pulley broke the language barrier and lured her to Cairo where she waited for
nine months, vainly hoping to see Farouk.
\end{quote}
In our system, {\it break} has one sense which means {\it loss of
functionality}. Its selection restriction is that the patient should
be a mechanical device which fails to match {\it language barrier}.
However, in our ontology, a {\it
language barrier} is supposed to be an entity having functionality
which has been placed in the nominal hierachy near the concept of
mechanical-device.
So the system can choose the {\it break} sense  {\it loss of
functionality} over all the other {\it break} senses
as the most probable one.  Based on this interpretation,
the system can correctly select the Chinese verb
\psfig{figure=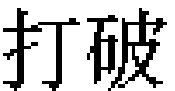,height=8pt}
{\it da-po} as the target realization. The correct selection becomes
possible because the system has a measurement for the degree of  satisfaction
of the selection restrictions.
In another example,
\begin{quote}\it
ca43:10:Other tax-exempt bonds of State and local governments hit
a price peak on February 21, according to Standard \& Poor 's
average.
\end{quote}
{\it hit} is defined with the concepts
{\it \%move-toward-in-space \%contact-in-space \%receive-force}.
Since {\it tax-exempt bonds} and {\it a price peak} are not physical
objects, the argument structure is excluded from the HIT usage type.
If the system has the knowledge that
price can be changed in value and fixed at some value, and these
concepts of  {\it
change-in-value} and {\it fix-at-value} are near the concepts {\it
\%move-toward-in-space \%contact-in-space}, the system can
interpret the meaning as {\it change-in-value} and {\it fix-at-value}.
In this case, the correct lexical selection can be made as
\psfig{figure=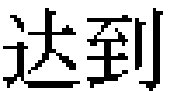,height=8pt} {\it
da-dao}.  This result is predicated on the definition of {\it hit}
as having concepts in three domains that are all structurally related, i.e.,
nearby in the hierarchy, the concepts related to prices.


\section*{Methodology and experimental results}
\vspace*{-0.1cm}
Our UNICON system translates a subset (the more concrete
usages) of the English {\it break} verbs from the Brown corpus
into Chinese with larger freedom to choose the target verbs and more
accuracy than
the TranStar system.
Our coverage has been extended to include
verbs from the  semantically similar
{\it hit, touch, break} and {\it cut} classes as defined by Beth
Levin. Twenty-one English verbs from these classes have been encoded
in the system. Four hundred  Brown corpus sentences which contain these 21
English verbs have been selected, Among them, 100 sentences with
concrete objects are used as training samples. The verbs were translated
into Chinese verbs. The other 300 sentences are divided into two
test sets. Test set one contains 154
sentences that are carefully chosen to make sure
the verb takes a concrete object as its patient.
For test set one, the lexical selection of the system got a correct
rate 57.8\% before encoding
the meaning of the unknown verb arguments; and a correct rate 99.45\% after
giving the unknown English words conceptual meanings in the system's
conceptual hierarchy. The second test set contains
 116 sentences including sentences
with non-concrete objects, metaphors, etc.
The lexical selection of the system got a correct rate of 31\% before encoding
the unknown verb arguments, a 75\% correct rate after adding meanings
and a 88.8\% correct rate after extended selection process applied. The
extended selection process relaxes the constraints and attempts to
find out the best possible target verb with the similarity measure.

{}From these tests, we can see the benefit of defining the verbs on
several cognitive domains. The conceptual hierarchical structure
provides a way of measuring the similarities among different verb
senses; with relaxation, metaphorical processing becomes possible.
The correct rate is improved by 13.8\% by using this extended selection
process.


\section*{Discussion}
With examples from the translation of English to Chinese
we have shown that
verb semantic representation has great impact on the quality of
lexical selection. Selection restrictions on verb arguments can only
define  default situations for verb events,
and are often overridden
by context information. Therefore, we propose a novel method
for defining
verbs based on a set of shared semantic domains. This representation
scheme not only takes  care of the semantic-syntactic correspondence,
but also provides similarity measures for the system
for the performance of inexact
matches based on verb meanings. The conceptual similarity has
priority over selection constrants on the verb arguments.
We leave scaling up the system to future work.

\small
\nocite{dorr:lexical,nirenburg:machine,jackendoff:semantic}
\bibliography{fuzzy,lexical,theory,corpus}
\bibliographystyle{myapa}

\end{document}